\documentclass[runningheads, envcountsame, a4paper]{llncs}
\usepackage{booktabs}
\usepackage{graphicx}
\usepackage{lipsum}
\usepackage{subcaption}
\usepackage{multirow}
\usepackage{mathtools}
\usepackage{tabularx}
\usepackage{booktabs}
\usepackage{amsmath}
\usepackage{float}
\usepackage[misc]{ifsym}
\usepackage[hyphens]{url}
\usepackage{microtype}

\newcommand*\samethanks[1][\value{footnote}]{\footnotemark[#1]}
\begin{document}
\title{REXUP: I REason, I EXtract, I UPdate with Structured Compositional Reasoning \\for Visual Question Answering}
\titlerunning{REXUP: I REason, I EXtract, I UPdate for Visual Question Answering}

\toctitle{REXUP: I REason, I EXtract, I UPdate with Structured Compositional Reasoning for Visual Question Answering}

\author{Siwen Luo\thanks{Both authors are first author} \and
Soyeon Caren Han\samethanks{\Letter} \and
Kaiyuan Sun \and
Josiah Poon
}
\authorrunning{S. Luo et al.}
\tocauthor{Siwen~Luo, Soyeon~Caren~Han, Kaiyuan~Sun, Josiah~Poon}

\institute{School of Computer Science, 
The University of Sydney, 1 Cleveland Street, \\
The University of Sydney, NSW 2006, Australia, \\
\email{$\{$siwen.luo, caren.han, kaiyuan.sun, josiah.poon$\}$@sydney.edu.au}}

\maketitle              % typeset the header of the contribution
\begin{abstract}
Visual Question Answering (VQA) is a challenging multi-modal task that requires not only the semantic understanding of images and questions, but also the sound perception of a step-by-step reasoning process that would lead to the correct answer. So far, most successful attempts in VQA have been focused on only one aspect; either the interaction of visual pixel features of images and word features of questions, or the reasoning process of answering the question of an image with simple objects. In this paper, we propose a deep reasoning VQA model (REXUP- REason, EXtract, and UPdate) with explicit visual structure-aware textual information, and it works well in capturing step-by-step reasoning process and detecting complex object-relationships in photo-realistic images. REXUP consists of two branches, image object-oriented and scene graph-oriented, which jointly works with the super-diagonal fusion compositional attention networks. We evaluate REXUP on the benchmark GQA dataset and conduct extensive ablation studies to explore the reasons behind REXUP’s effectiveness. Our best model significantly outperforms the previous state-of-the-art, which delivers 92.7\% on the validation set, and 73.1\% on the test-dev set. 

\end{abstract}
\section{Introduction}
Vision-and-language reasoning requires the understanding and integration of visual contents and language semantics and cross-modal alignments. Visual Question Answering (VQA)~\cite{antol2015vqa} is a popular vision-and-language reasoning task, which requires the model to predict correct answers to given natural language questions based on their corresponding images. Substantial past works proposed VQA models that focused on analysing objects in photo-realistic images but worked well only for simple object detection and yes/no questions \cite{lu2016hierarchical,yu2017multi,kim2016hadamard}. To overcome this simple nature and improve the reasoning abilities of VQA models, the Clever dataset\cite{johnson2017clevr} was introduced with compositional questions and synthetic images, and several  models~\cite{hu2018explainable,perez2018film} were proposed and focused on models' inferential abilities. The state-of-the-art model on the Clevr dataset is the compositional attention network(CAN)\cite{hudson2018compositional}, which generates reasoning steps attending over both images and language-based question words. However, the Clevr dataset is specifically designed to evaluate reasoning capabilities of a VQA model. Objects in the Clevr dataset images are only in three different shapes and four different spatial relationships, which results in simple image patterns. Therefore, a high accuracy on Clevr dataset hardly prove a high object detection and analysis abilities in photo-realistic images, nor the distinguished reasoning abilities of a VQA model. To overcome the limitations of VQA and Clevr~\cite{antol2015vqa,goyal2017making}, the GQA dataset~\cite{hudson2019gqa} includes photo-realistic images with over 1.7K different kinds of objects and 300 relationships. GQA provides diverse types of answers for open-ended questions to prevent models from memorizing answer patterns and examine the understanding of both images and questions for answer prediction. 

The state-of-the-art models on the Clevr and VQA dataset suffered large performance reductions when evaluated on GQA \cite{hudson2018compositional,norcliffe2018learning,anderson2018bottom}. Most VQA works focus on the interaction between visual pixel features extracted from images and question features while such interaction does not reflect the underlying structural relationships between objects in images. Hence, the complex relationships between objects in real images are hard to learn.Inspired by this motivation, we proposed REXUP(REason, EXtract, UPdate) network to capture step-by-step reasoning process and detect the complex object-relationships in photo-realistic images with the scene graph features. A scene graph is a graph representation of objects, attributes of objects and relationships between objects where objects that have relations are connected via edge in the graph.

The REXUP network consists of two parallel branches where the image object features and scene graph features are respectively guided by questions in an iterative manner, constructing a sequence of reasoning steps with REXUP cells for answer prediction. A super-diagonal fusion is also introduced for a stronger interaction between object features and question embeddings. The branch that processes scene graph features captures the underlying structural relationship of objects, and will be integrated with the features processed in another branch for final answer prediction. Our model is evaluated on the GQA dataset and we used the official GQA scene graph annotations during training. To encode the scene graph features, we extracted the textual information from the scene graph and used Glove embeddings to encode the extracted textual words in order to capture the semantic information contained in the scene graph. In the experiments, our REXUP network achieved the state-of-the-art performance on the GQA dataset with complex photo realistic images in deep reasoning question answering task.

\section{Related Work and Contribution}
We explore research trends in diverse visual question answering models, including fusion-based, computational attention-based, and graph-based VQA models. 

\noindent\textbf{Fusion-based VQA} Fusion is a common technique applied in many VQA works to integrate language and image features into a joint embedding for answer prediction. There are various types of fusion strategies for multi-modalities including simple concatenation and summation. For example, \cite{shrestha2019answer} concatenated question and object features together and pass the joint vectors to a bidirectional GRU for further processes. However, the recent bilinear fusion methods are more effective at capturing higher level of interactions between different modalities and have less parameters in calculation. For example, based on the tensor decomposition proposed in~\cite{ben2017mutan}, \cite{ben2019block} proposed a block-term decomposition of the projection tensor in bilinear fusion. \cite{cadene2019murel} applied this block-term fusion in their proposed MuRel networks, where sequences of MuRel cells are stacked together to fuse visual features and question features together. 

\noindent\textbf{Computational Attention-based VQA} Apart from fusion techniques, attention mechanisms are also commonly applied in VQA for the integration of multi-modal features. Such attention mechanisms include soft attention mechanism like ~\cite{hudson2018compositional,anderson2018bottom} using softmax to generate attention weights over object regions and question words, self attention mechanism like~\cite{yu2019deep,nguyen2018improved} that applied dot products on features of each mode, and co-attention mechanisms like in~\cite{liu2019densely,Gao_2019_CVPR} using linguistic features to guide attentions of visual features or vice versa. 

\noindent\textbf{Graph Representations in VQA} In recent years, more works have been proposed to integrate graph representations of images in VQA model. \cite{norcliffe2018learning} proposed a question specific graph-based model where objects are identified and connected with each other if their relationships are implied in the given question. There are also works use scene graph in VQA like we did. \cite{shi2019explainable} integrates scene graphs together with functional programs for explainable reasoning steps. \cite{haurilet2019s} claimed only partial image scene graphs are effective for answer prediction and proposed a selective system to choose the most important path in a scene graph and use the most probable destination node features to predict an answer. However, these works did not apply their models on GQA. 
\paragraph{}
\noindent\textbf{REXUP's Contribution} In this work, we move away from the classical attention and traditional fusion network, which have been widely used in simple photo-realistic VQA tasks and focus mainly on the interaction between visual pixel features from an image and question embeddings. Instead, we focus on proposing a deeper reasoning solution in visual-and-language analysis, as well as complex object-relationship detection in complex photo-realistic images. We propose a new deep reasoning VQA model that can be worked well on complex images by processing both image objects features and scene graph features and integrating those with super-diagonal fusion compositional attention networks.

\section{Methodology}
\begin{figure*}[tbp]
    \centering
    \begin{subfigure}[b]{0.28\textwidth}
        \centering
        \includegraphics[width=3.3cm]{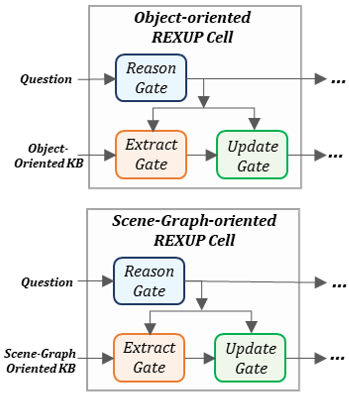}
        \caption{REXUP Network}
        \label{fig:1a}
    \end{subfigure}
    \begin{subfigure}[b]{0.68\textwidth}
        \centering
        \includegraphics[width=8.2cm]{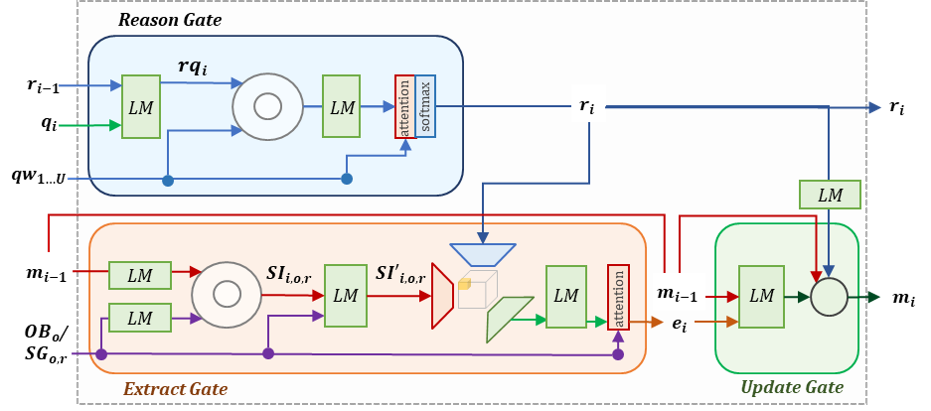}
        \caption{REXUP Cells}
        \label{fig:1b}
    \end{subfigure}
    \caption{\textbf{REXUP Network and REXUP cell.} (a) The REXUP network includes two parallel branches, object-oriented \textit{(top)} and scene graph-oriented \textit{(bottom)}. (b) A REXUP cell contains reason, extract, and update gate which conduct multiple compositional reasoning and super-diagonal fusion process}
    \label{Fig3}
\end{figure*}

The REXUP network contains two parallel branches, object-oriented branch and scene-graph oriented branch, shown in Fig.~\ref{fig:1a}. Each branch contains a sequence of REXUP cells where each cell operates for one reasoning step for the answer prediction. As shown in Fig.~\ref{fig:1b}, each REXUP cell includes a reason, an extract and an update gate. At each reasoning step, the reason gate identifies the most important words in the question and generates a current reasoning state with distributed attention weights over each word in the question. This reasoning state is fed into the extract gate and guides to capture the important objects in the knowledge base, retrieving information that contains the distributed attention weights over objects in the knowledge base. The update gate takes the reasoning state and information from extract gate to generate the current memory state.

% Reason gate and memory gate are the same for cells in both SGKB branch and OKB branch. But in extract gate, SGKB branch takes scene graph features as inputs while the OKB branch takes image object features as inputs and super-diagonal fusion is replaced with element-wise multiplication.

\subsection{Input Representation}
Both Object-oriented branch and Scene graph-oriented branch take question and knowledge base as inputs; image object-oriented knowledge base (OKB) and scene-graph-oriented knowledge base (SGKB). For a question $q\in Q$ with maximum $U$ words, contextual words are encoded via a pre-trained $300d$ Glove embedding and passed into bi-LSTM to generate a sequence of hidden states $qw_{1...U}$ with $d$ dimension for question contextual words representation. The question is encoded by the concatenation of the last backward and forward hidden states, $\overleftarrow{qw_{1}}$ and $\overrightarrow{qw_{U}}$. Object features are extracted from a pre-trained Fast-RCNN model, each image contains at most 100 regions represented by a $2048d$ object feature. For each $o_{th}$ object in an image, linear transformation converts the object features with its corresponding coordinates to a $512d$ object region embedding. The SGKB is the matrix of scene graph objects each of which is in 900 dimensions after concatenating with their corresponding attribute and relation features. To encode the scene graph object features, all the objects names, their attributes and relations in the scene graph are initialized as $300d$ Glove embedding. For each object's attributes, we take the average of those attributes features $A$. For each object's relations, we first average each relation feature $r_{s}\in R$ and the subject feature $o_{j}\in O$ that it is linked to, and then average all such relation-subject features that this object $o_{n}\in O$ has as its final relation feature. We concatenate the object feature, attribute feature and relation feature together as one scene graph object feature $SG_{o,r}$ of the whole scene graph.

\subsection{REXUP Cell}
With the processed input, each branch consists of a sequence of REXUP cells where each cell operates for one reasoning step for the answer prediction.

\subsubsection{Reason Gate} At each reasoning step, the reason gate in each REXUP cell $i=1,...,P$ takes the question feature $q$, the sequence of question words $qw_{1}, qw_{2},..., qw_{U}$ and the previous reasoning state $r_{i-1}$ as inputs. Before being passed to the reason gate, each question $q$ is processed through a linear transformation $q_{i}=W_{i}^{d\times 2d}q+b_{i}^{d}$ to encode the positional-aware question embedding $q_{i}$ with $d$ dimension in the current cell. A linear transformation is then processed on the concatenation of $q_{i}$ and the previous reasoning state $r_{i-1}$,
\begin{equation}
    rq_{i}=W^{d\times 2d}\left [ r_{i-1}, q_{i} \right ]+b^{d}
\end{equation}
in order to integrate the attended information at the previous reasoning step into the question embedding at the current reasoning step.

Then an element-wise multiplication between $rq_{i}$ and each question word $qw_{u}$, where $u=1,2,...,U$, is conducted to transfer the information in previous reasoning state into each question word at the current reasoning step, the result of which will be processed through a linear transformation, yielding a sequence of new question word representations $ra_{i,1},...,ra_{i,u}$ containing the information obtained in previous reasoning state. A softmax is then applied to yield the distribution of attention scores $rv_{i,1},...,rv_{i,u}$ over question words $qw_{1},...,qw_{u}$.
\begin{equation}
      ra_{i,u}=W^{1\times d}(rq_{i}\odot qw_{u})+b
\end{equation}
\begin{equation}
    rv_{i,u}=softmax(ra_{i,u})
\end{equation}
\begin{equation}
    r_{i}=\sum_{u=1}^{U}rv_{i,u}\cdot qw_{u}
\end{equation}
The multiplications of each $rv_{i,u}$ and question word $qw_{u}$ are summed together and generates the current reasoning state $r_{i}$ that implies the attended information of a question at current reasoning step.

\subsubsection{Extract Gate}
The extract gate takes the current reasoning state $r_{i}$, previous memory state $m_{i-1}$ and the knowledge base features as inputs. For the OKB branch, knowledge base features are the object region features $OB_{o}$, and for the SGKB branch, knowledge base features are the scene graph features $SG_{o,r}$. For each object in the knowledge base, we first multiplied its feature representation with the previous memory state to integrate the memorized information at the previous reasoning step into the knowledge base at the current reasoning step, the result of which is then concatenated with the input knowledge base features and projected into $d$ dimensions by a linear transformation. This interaction $SI_{i,o,r}'$ generates the knowledge base features that contains the attended information memorized at the previous reasoning step as well as the yet unattended information of knowledge base at current reasoning step. The process of the extract gate in the SGKB branch can be shown in the following equations, where the interaction $SI_{i,o,r}'$ contains the semantic information extracted from the object-oriented scene graph.
\begin{equation} \label{eq:5}
    SI_{i,o,r}=\left [ W_{m}^{d\times d}m_{i-1}+b_{m}^{d} \right ]\odot \left [ W_{s}^{d\times d}SG_{o,r}+b_{s}^{d} \right ]
\end{equation}
We then make $SI_{i,o,r}'$ interact with $r_{i}$ to let the attended question words guide the extract gate to detect important objects of knowledge base at the current reasoning step. In the SGKB branch, such integration is completed through a simple multiplication as shown in \eqref{eq:7}.
\begin{equation} \label{eq:6}
    SI_{i,o,r}'=W^{d\times 2d}\left [ SI_{i,o,r}, SG_{o,r} \right ]+b^{d}
\end{equation}
\begin{equation} \label{eq:7}
    ea_{i,o,r}=W^{d\times d}(r_{i}\odot SI_{i,o,r}')+b^{d}
\end{equation}
However, in OKB branch, $SG_{o,r}$ in Equation \eqref{eq:5} and \eqref{eq:6} is replaced with the object region features $OB_{o}$, and generated interaction $I_{i,o}'$, which will be integrated with $r_{i}$ through a super-diagonal fusion~\cite{ben2019block} as stated in Equation (8), where $\theta$ is a parameter to be trained. Super-diagonal fusion projects two vectors into one vector with $d$ dimension through a projection tensor that would be decomposed into three different matrices during calculation in order to decrease the computational costs while boosting a stronger interaction between input vectors. The resulted $F_{r_{i},I_{i,o}'}$ is passed via a linear transformation to generate $ea_{i,o}$.
\begin{equation}
F_{r_{i},I_{i,o}'}=SD(r_{i}, I_{i,o}'; \theta) \quad \text{and} \quad  ea_{i,o}=W^{d\times d}F_{r_{i},I_{i,o}'}+b^{d}  \tag*{(8) and (9)}\label{8and9}
\end{equation}
Similar to the process in the reason gate, $ea_{i,o,r}$ and $ea_{i,o}$ are then processed by softmax to get the distribution of attention weights for each object in the knowledge base. The multiplications of each $ea_{i,o,r}$/$ea_{i,o}$ and knowledge base $SG_{o,r}$/$OB_{o}$ are summed together to yield the extracted information $e_{i}$. 
\begin{equation}
    ev_{i,o,r}=softmax(ea_{i,o,r}) \quad\text{ and }\quad ev_{i,o}=softmax(ea_{i,o}) \tag*{(10)}
\end{equation}
\begin{equation}
    e_{i}=\sum_{o=1}^{O}ev_{i,o,r}\cdot SG_{o,r} \quad\text{ and }\quad e_{i}=\sum_{o=1}^{O}ev_{i,o}\cdot OB_{o} \tag*{(11)}
\end{equation}

\subsubsection{Update Gate}
We apply a linear transformation to the concatenation of the extracted information $e_{i}$ and previous memory state $m_{i-1}$ to get $m_{i}^{prev}$.
\begin{equation}\label{eq:12}
    m_{i}^{prev}=W^{d\times 2d}\left [ e_{i}, m_{i-1} \right ]+b^{d}\tag*{(12)}
\end{equation}
\begin{equation}\label{eq:13}
    m_{i}=\sigma (r'_{i})m_{i-1}+(1-\sigma (r'_{i}))m'_{i}\tag*{(13)}
\end{equation}
To reduce redundant reasoning steps for short questions, we applied sigmoid function upon $m_{i}^{prev}$ and $r'_{i}$, where $r'_{i}=W^{1 \times d}r_{i}+b^{1}$, to generate the final memory state $m_{i}$.

The final memory states generated in the OKB branch and SGKB branch respectively are concatenated together as the ultimate memory state $m_{P}$ for overall $P$ reasoning steps. $m_{P}$ is then integrated with the question sentence embedding $q$ for answer prediction. In this work, we set $P=4$. 

\section{Evaluation}
\subsection{Evaluation Setup}
\subsubsection{Dataset} Our main research aim is proposing a new VQA model that provides not only complex object-relationship detection capability, but also deep reasoning ability. Hence, we chose the GQA that covers 1) complex object-relationship: 113,018 photo-realistic images and 22,669,678 questions of five different types, including \textit{Choose, Logical, Compare, Verify and Query}, and 2) deep reasoning tasks: over 85\% of questions with 2 or 3 reasoning steps and 8\% of questions with 4+ reasoning steps. The GQA is also annotated with scene graphs extracted from the Visual Genome~\cite{krishna2017visual} and functional programs that specify reasoning operations for each pair of image and question. The dataset is split into 70\% training, 10\% validation, 10\% test-dev and 10\% test set. 

\subsubsection{Training Details} The model is trained on GQA training set for 25 epochs using a 24 GB NVIDIA TITAN RTX GPU with 10.2 CUDA toolkit. The average per-epoch training times and total training times are 7377.31 seconds and 51.23 hours respectively. We set the batch size to 128 and used an Adam optimizer with an initial learning rate of 0.0003.

\subsection{Performance Comparison} 
In Table.~\ref{tab1}, we compare our model to the state-of-the-art models on the validation and test-dev sets of GQA. Since the GQA test-dev set does not provide pre-annotated scene graphs, we used the method proposed in~\cite{zellers2018neural} to predict relationships between objects and generate scene graphs from images of GQA test-dev set for the evaluation procedure. However, the quality of the generated scene graphs are not as good as the pre-annotated scene graphs in the GQA validation set, which lead to the decreased performance on test-dev. Nevertheless, our model still achieves the state-of-the-art performance with 92.7\% on validation and 73.1\% on test-dev. Compared to \cite{anderson2018bottom,hudson2018compositional,tan2019lxmert} that only used the integration between visual pixel features and question embedding through attention mechanism, our model applies super-diagonal fusion for a stronger interaction and also integrates the scene graph features with question embedding, which help to yield much higher performance. Moreover, our model greatly improves over \cite{hu2019language}, which used the graph representation of objects but concatenated the object features with contextual relational features of objects as the visual features to be integrated with question features through the soft attention. The significant improvement over \cite{hu2019language} indicates that the parallel training of OKB and SGKB branch can successfully capture the structural relationships of objects in images. 

\begin{table}
\centering
\caption{State-of-the-art performance comparison on the GQA dataset}\label{tab1}
\begin{tabularx}{1\textwidth} { 
   |>{\centering\arraybackslash\hsize=.5\hsize}X 
   |>{\centering\arraybackslash\hsize=.25\hsize}X 
   |>{\centering\arraybackslash\hsize=.25\hsize}X| }
 \hline 
 {\bfseries Methods} & {\bfseries Val} & {\bfseries Test-dev} \\
 \hline
 CNN+LSTM~\cite{hudson2019gqa} & 49.2  & -  \\ \hline
 Bottom-Up~\cite{anderson2018bottom} & 52.2  & -   \\ \hline
 MAC~\cite{hudson2018compositional}  & 57.5  & -  \\ \hline
 LXMERT~\cite{tan2019lxmert}  & 59.8  & 60.0  \\ \hline
 single-hop~\cite{hu2019language}  & 62  & 53.8   \\ \hline
 single-hop+LCGN~\cite{hu2019language}  & 63.9  & 55.8 \\ \hline
 \textbf{Our Model}  & \textbf{92.7}  & \textbf{73.1}  \\
 \hline
\end{tabularx}
\end{table}

\begin{table}
\centering
\caption{Results of ablation study on \textbf{validation} and \textbf{test-dev} set of GQA. `O' and `X' refers to the existence and absence of scene-graph oriented knowledge branch(\textit{SGKB}) and super-diagonal(\textit{SD}) fusion applied in object-oriented knowledge branch(\textit{OKB}) branch respectively}\label{tab2}
\begin{tabularx}{1\textwidth} {
  |>{\centering\arraybackslash\hsize=.2\hsize}X
  |>{\centering\arraybackslash\hsize=.4\hsize}X 
  |>{\centering\arraybackslash\hsize=.4\hsize}X
  |>{\centering\arraybackslash\hsize=.55\hsize}X
  |>{\centering\arraybackslash\hsize=.6\hsize}X
  |>{\centering\arraybackslash\hsize=0.8\hsize}X| }
 \hline
 {\bfseries \#} & {\bfseries OKB} & {\bfseries SD} & {\bfseries SGKB} & {\bfseries Val} & {\bfseries Test-dev} \\
 \hline
 1 & O & X & X  & 62.35  & 56.92  \\ \hline
 2 & O & O & X  & 63.10  & 57.25  \\ \hline
 3 & O & X & O  & 90.14  & 72.38  \\ \hline
 \textbf{4} & \textbf{O} & \textbf{O} & \textbf{O}  & \textbf{92.75}  & \textbf{73.18}  \\
 \hline
\end{tabularx}
\end{table}

\subsection{Ablation study} 
We conducted the ablation study to examine the contribution of each component in our model. As shown in Table~\ref{tab2}, integrating object-oriented scene graph features is critical in achieving a better performance on the GQA. Using only OKB branch leads to a significant drop of 29.65\% in the validation accuracy and 15.93\% in the test-dev accuracy. The significant performance decrease also proves the importance of semantic information of objects' structural relationships in VQA tasks. Moreover, applying the super-diagonal fusion is another key reason of our model's good performance on GQA. We compared performances of models that apply super-diagonal fusion and models that apply element-wise multiplication. The results show that using element-wise multiplication causes a drop of 2.61\% on the validation set and 0.8\% on the test-dev set. It still shows that the concrete interaction between image features and question features generated by super-diagonal fusion contributes to an improved performance on the GQA.

\begin{table}
\centering

\caption{Parameter Testing with different number of the REXUP cell}\label{tab3}
\begin{tabularx}{1\textwidth} { 
   |>{\centering\arraybackslash}X 
   |>{\centering\arraybackslash}X 
   |>{\centering\arraybackslash}X| }
 \hline
 {\bfseries \# of cells} & {\bfseries Val} & {\bfseries Test-dev} \\
 \hline
 1 & 90.97 & 72.08   \\ \hline
 2 & 90.98 & 72.13  \\ \hline
 3 & 92.67 & 72.56  \\ \hline
 \textbf{4} & \textbf{92.75} & \textbf{73.18}  \\
 \hline
\end{tabularx}
\end{table}

\begin{figure}[tb]
  \centering\includegraphics[width=9.5cm]{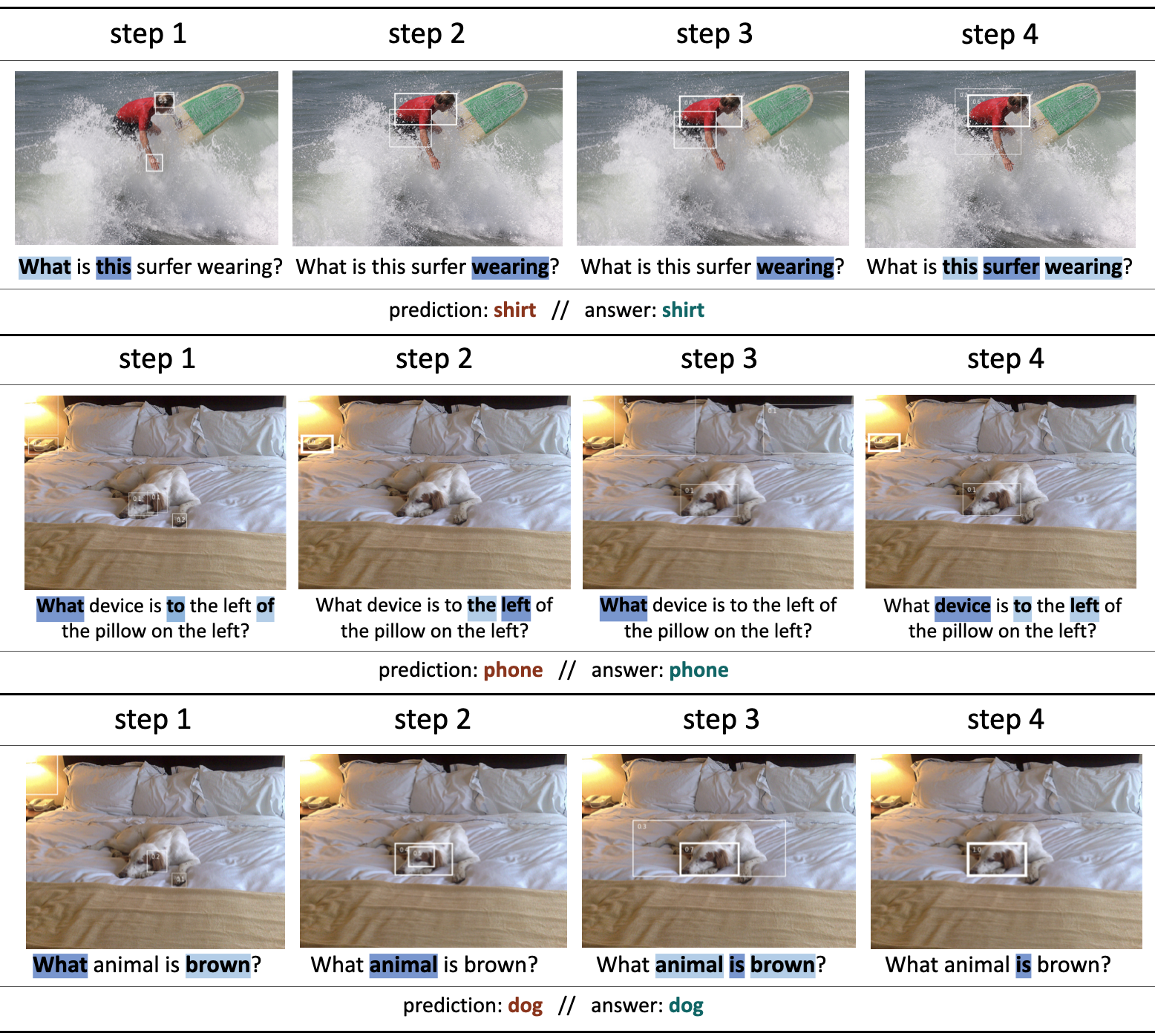}
  \caption{Visualization of important image objects and question words at each reasoning step. Object regions with high attention weights are framed with white bounding boxes. The thicker the frame, the more important the object region is. Question words with high attention weights are colored blue in the question.}
  \label{Figure1}
\end{figure}

\begin{figure*}
    \centering
    \begin{subfigure}[b]{0.45\textwidth}
        \centering
        \includegraphics[width=5cm]{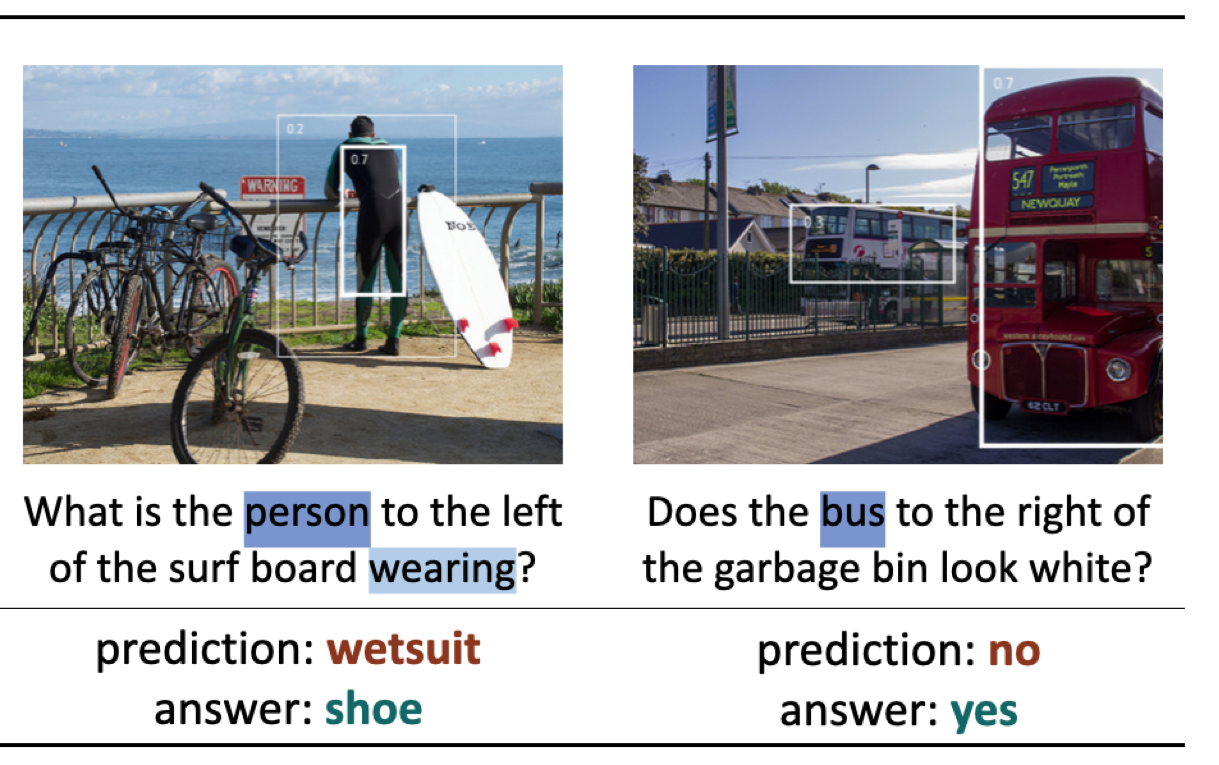}
        \caption{}
        \label{fig:a}
    \end{subfigure}
    \begin{subfigure}[b]{0.45\textwidth}
        \centering
        \includegraphics[width=5cm]{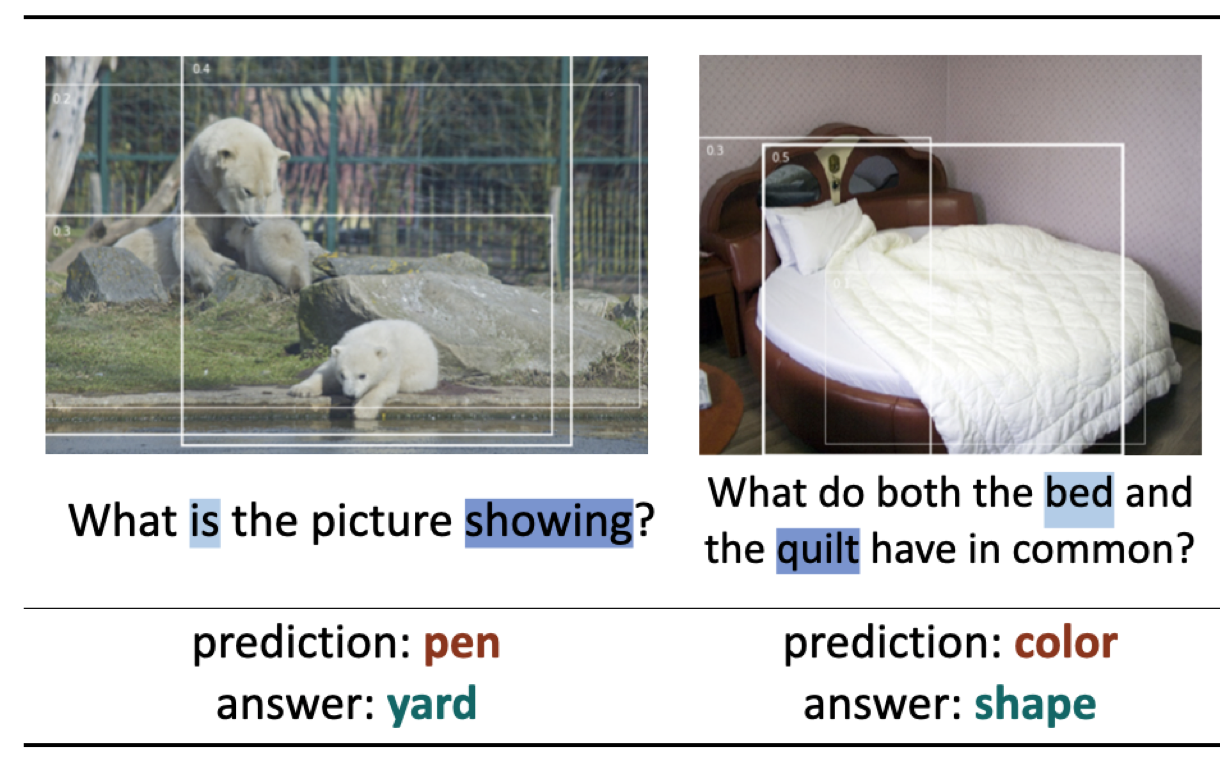}
        \caption{}
        \label{fig:b}
    \end{subfigure}
    \caption{Figure \ref{fig:a} shows examples when the ground truth answer and our prediction are both correct to the given question. Figure \ref{fig:b} shows examples when our prediction is more accurate than the ground truth answer in dataset}
    \label{Fig3ab}
\end{figure*}

\subsection{Parameter Comparison}
Sequences of REXUP cells will lead to sequential reasoning steps for the final answer prediction. The three gates in each cell are designed to follow questions' compositional structures and retrieve question-relevant information from knowledge bases at each step. To reach the ultimate answer, a few reasoning steps should be taken, and less cells are insufficient to extract the relevant knowledge base information for accurate answer prediction, especially for compositional questions with longer length. In order to verify this assumption, we have conducted experiments to examine the model's performances with different numbers of REXUP cells in both branches. The results of different performances are shown in Table~\ref{tab3}. From the result, we can see that the prediction accuracy on both validation and test-dev set will gradually increase (90.97\% to 92.75\% on validation and 72.08\% to 73.18\% on test-dev) as the cell number increases. After experiment, we conclude that four REXUP cells are best both for clear presentation of reasoning capabilities and a good performance on the GQA.

\subsection{Interpretation}
To have a better insight into the reasoning abilities of our model, we extract the linguistic and visual attention weights our model computes at each reasoning step to visualize corresponding reasoning processes. Taking the first row in Fig.~\ref{Figure1} as an example, at the first reasoning step, concrete objects - man's hand and head obtain high visual attention score. When it comes to the second and third reasoning step, linguistic attention focuses on \textit{wearing} and corresponding visual attention focuses on man's shirt and pants. This indicates that our model's abilities in capturing the underlying semantic words of questions as well as detecting relevant objects in image for answer prediction. Moreover, our model's good understanding of both images and questions is also shown when given different questions for a same image. For example, in the second row in Fig.~\ref{Figure1}, the model successfully captures the \textbf{\textit{phone}} in image for the question, but for images of third row in Fig.~\ref{Figure1}, the \textbf{\textit{dog}} is detected instead. We also found that sometimes our predicted answer is correct even though it's different from the answer in dataset. For example, in the first image of Fig. \ref{fig:a}, our model assigns a high visual attention score to wetsuit in image when question words \textit{person} and \textit{wearing} are attended. Our model then gives the prediction \textbf{\textit{wetsuit}}, which is as correct as \textbf{\textit{shoe}} considering the given image and question. Similarly, in the second image of Fig.~\ref{fig:a}, both white bus and red bus are spatially on the right of garbage. Our model captures both buses but assigns more attention to the red bus that is more obvious on the picture and predicts \textit{no}, which is also a correct answer to the question. In addition, we found that in some cases our model's answer is comparatively more accurate than the annotated answer in dataset. For example, for first image of Fig.~\ref{fig:b}, \textbf{\textit{pen}}, as a small area surrounded by fence to keep animal inside, is more accurate than the annotated answer \textbf{\textit{yard}}. Likewise, the bed and quilt are actually different in shape but both in white color, which makes our model's answer correct and the ground truth answer incorrect.

\section{Conclusion}
In conclusion, our REXUP network worked well in both capturing step-by-step reasoning process and detecting a complex object-relationship in photo-realistic images. Our proposed model has achieved the state-of-the-art performance on the GQA dataset, which proves the importance of structural and compositional relationships of objects in VQA tasks. Extracting the semantic information of scene graphs and encoding them via textual embeddings are efficient for the model to capture such structural relationships of objects. The parallel training of two branches with object region and scene graph features respectively help the model to develop comprehensive understanding of both images and questions.

%\bibliographystyle{splncs04}
%\bibliography{mybibliography}
%

\end{document}